\pdfoutput=1

\documentclass[11pt]{article}

\usepackage[final]{acl}

\usepackage{booktabs}
\usepackage{tcolorbox}
\usepackage{tabularx}

\usepackage{pifont} 
\usepackage{colortbl}
\usepackage{xcolor}
\usepackage{amssymb}
\usepackage{amsmath}
\usepackage{multirow}

\usepackage{times}
\usepackage{latexsym}

\usepackage[T1]{fontenc}

\usepackage[utf8]{inputenc}

\usepackage{microtype}

\usepackage{inconsolata}

\usepackage{graphicx}

\title{Examining Multilingual Embedding Models Cross-Lingually Through LLM-Generated Adversarial Examples}

\author{
\parbox{0.6\linewidth}{\centering
Andrianos Michail ~~ Simon Clematide ~~ Rico Sennrich}
\\
University of Zurich
\\
\texttt{\{andrianos.michail,simon.clematide,sennrich\}@cl.uzh.ch}}

\begin{document}
\maketitle
\begin{abstract}
The evaluation of cross-lingual semantic search models is often limited to existing datasets from tasks such as information retrieval and semantic textual similarity. We introduce Cross-Lingual Semantic Discrimination (CLSD), a lightweight evaluation task that requires only parallel sentences and a Large Language Model (LLM) to generate adversarial distractors. CLSD measures an embedding model's ability to rank the true parallel sentence above semantically misleading but lexically similar alternatives. As a case study, we construct CLSD datasets for German--French in the news domain. Our experiments show that models fine-tuned for retrieval tasks benefit from pivoting through English, whereas bitext mining models perform best in direct cross-lingual settings. A fine-grained similarity analysis further reveals that embedding models differ in their sensitivity to linguistic perturbations.\footnote{We release our code and datasets under AGPL-3.0: \url{https://github.com/impresso/cross_lingual_semantic_discrimination}}
\end{abstract}

\section{Introduction}

When selecting a model for semantic search in a specific domain and language pair, standard embedding benchmarks such as (Multilingual) MTEB \citep{muennighoff-etal-2023-mteb, enevoldsenmmteb} provide only partial coverage. We argue that evaluation should be carried out directly on the target language pair and domain of interest. However, the lack of suitable datasets makes it difficult to reach well-founded decisions about applications, such as choosing an embedding model or deciding whether to use English as a pivot language.

We propose a simple and efficient method for dataset generation and evaluation that simulates real-world challenges of cross-lingual semantic retrieval in large text collections. Our approach uses LLMs to create adversarial examples from parallel sentences, designed to test the cross-lingual embedding capabilities of multilingual models. Concretely speaking, four distractor sentences are generated for each parallel target sentence. We require these sentences to be very similar to the original target sentence in terms of syntactic structure and surface word forms, but to be semantically dissimilar.  
Our proposed task, Cross-Lingual Semantic Discrimination (CLSD), involves identifying the correct target sentence from among four distractor sentences in the target language, given an original sentence in the source language.

To better understand how these distractor modifications influence semantic similarity both cross-lingually and monolingually, we perform a fine-grained analysis and correlate the semantic changes with linguistic properties at the level of part-of-speech tags.

\paragraph{\textbf{Contributions}}
\begin{itemize}
    \item We propose a new task that emulates cross-lingual semantic search within large text collections, requiring only a set of parallel sentences for dataset creation.
    \item We publish four such datasets in the news domain for the German--French language pair.
    \item We show that some embedding models perform better in direct cross-lingual retrieval, while others benefit from using English as a pivot language.
    \item We provide a linguistically informed analysis that offers insights into cross- and monolingual semantic representations of multilingual models.
\end{itemize}

\section{Related Work}

\begin{table*}[t]
    \centering \footnotesize
    \resizebox{.95\textwidth}{!}{%
    \begin{tabular}{p{0.325\textwidth}@{\hskip 0.05in}|p{0.305\textwidth}@{\hskip 0.05in}|p{0.305\textwidth}}
    \toprule
        \multicolumn{1}{c|}{\textbf{Original Sentence} $S$} &
        \multicolumn{1}{c|}{\textbf{Distractor Sentence 1}  $D_1$} & \multicolumn{1}{c}{\textbf{Distractor Sentence 2} $D_2$} \\ 
        \midrule
      Die Linkspartei beschließt in Bonn ihr Programm zur Europawahl. \newline\textit{(The Left Party adopts its program for the European elections in Bonn.)} & Die Linkspartei beschließt in Bonn ihr Programm zur \textcolor{red}{Bundestagswahl}. \newline\textit{(The Left Party adopts its program for the \textcolor{red}{Bundestag} elections in Bonn.)}& Die Linkspartei \textcolor{red}{verweigert} in Bonn ihr Programm zur Europawahl. \newline\textit{(The Left Party \textcolor{red}{refuses} its program for the European elections in Bonn.)}\\
      \midrule
      Là-bas, la perspective de la fin de la guerre reste toujours très éloignée. \newline\textit{(Over there, the prospect of the end of the war is still a long way off.)}& Là-bas, la fin de la guerre \textcolor{red}{semble} toujours très \textcolor{red}{incertaine}. \newline\textit{(Over there, the end of the war still \textcolor{red}{seems very uncertain}.)}& Là-bas, la perspective de la \textcolor{red}{paix semble} toujours très éloignée. \newline\textit{(Over there, the prospect of \textcolor{red}{peace} still \textcolor{red}{seems} a long way off.)}\\
      \bottomrule
    \end{tabular}
    }
    \caption{Examples of generated distractor sentences in German and French, with English translations provided in italics. Text in red indicates the modified segments.}
    \label{tab:transformation-examples}
\end{table*}

\paragraph{\textbf{Cross-Lingual Semantic Search}}

Cross-lingual semantic search is an umbrella term for tasks that involve searching for texts with similar or relevant semantics in another language. Examples of such cross-lingual tasks are information retrieval (CLIR) \citep{lawrie2023overviewtrec2022neuclir,lawrie2024overviewtrec2023neuclir}, question answering (CLQA) \citep{lewis-etal-2020-mlqa}, semantic text similarity (X-STS) \citep{cer2017semeval}, and bitext mining \citep{tatoeba, zweigenbaum2018overview}. As showcased in MMTEB \citep{enevoldsenmmteb}, multilingual embedding models vary in their performance for different cross-lingual semantic search tasks and language pairs. When selecting models for a use case, we often rely on results from other domains or language pairs, as direct evaluation results are not always available.

\paragraph{\textbf{Distractor Generation}}

PAWS-X \citep{yang2019paws,zhang-etal-2019-paws} generated adversarial examples by rule-based and backtranslation methods to measure the limitations of paraphrase identification models.  More recently, XSim++ \citep{chen-etal-2023-xsim} used rule-based text augmentation to generate synthetic examples for tuning LASER models (replacements by antonyms, swapping entities). Within semantic search, InPars \citep{bonifacio2022mmarcomultilingualversionms} relies solely on in-domain synthetic data to fine-tune retrieval models that surpass strong baselines. Lastly, \citet{li2025sentencesmithformallycontrollable} used AMR \citep{banarescu2013abstract} to craft hard negatives and examine the robustness of English embedding models.

\paragraph{\textbf{Bitext Mining}}

Bitext Mining is an established and most similar cross-lingual semantic search task to our proposed CLSD task. Bitext mining \citep{tatoeba, zweigenbaum2018overview} challenges embedding models to identify an exact parallel sentence from a large pool—often consisting of thousands of unrelated or loosely related sentences-originating from large, homogeneous corpora.

\begin{table*}[bt]

\resizebox{\textwidth}{!}{%
\begin{tabular}{c|ccc!{\vrule width 1pt}cc|cc!{\vrule width 1pt}c}
      \toprule
      \multirow{2}{*}{Model (Hugging Face Name)} 
      & \multicolumn{3}{c!{\vrule width 1pt}}{\textbf{Alignment}} 
      & \multicolumn{2}{c|}{\textbf{WMT19}} 
      & \multicolumn{2}{c!{\vrule width 1pt}}{\textbf{WMT21}} 
      & \multirow{2}{*}{Average} \\
      \cmidrule(lr){2-4} \cmidrule(lr){5-6} \cmidrule(lr){7-8}
      & \begin{tabular}{@{}c@{}}Cross-\\lingual\end{tabular}
      & \begin{tabular}{@{}c@{}}Para-\\phrase\end{tabular}
      & \begin{tabular}{@{}c@{}}Retrie-\\val\end{tabular}
      & DE$\rightarrow$FR & FR$\rightarrow$DE 
      & DE$\rightarrow$FR & FR$\rightarrow$DE 
      & \\
      \midrule
      \multicolumn{9}{c}{\textbf{Cross-Lingual CLSD Evaluation}} \\
      \hline
      multilingual-e5-base & \ding{51} & \ding{51} & \ding{51} & 91.51 & 88.46 & 86.34 & 81.97 & 87.07 \\
      multilingual-e5-large & \ding{51} & \ding{51} & \ding{51} & 94.50 & 91.51 & 91.38 & 87.57 & 91.24 \\
      gte-multilingual-base & \ding{51} & \ding{51} & \ding{51} & 90.22 & 89.68 & 90.48 & 91.60 & 90.50 \\
      paraphrase-multilingual-mpnet-base & \ding{51} & \ding{51} & \ding{55} & 91.31 & 91.11 & 91.15 & 92.95 & 91.63 \\
      sentence-transformers/LaBSE & \ding{51} & \ding{51} & \ding{55} & \textbf{95.18} & \textbf{94.30} & \textbf{94.06} & \textbf{94.18} & \textbf{94.43} \\
      \midrule
      \multicolumn{9}{c}{\textbf{English Pivot Translation by M2M 1.2B \citep{JMLR:v22:20-1307} CLSD Evaluation}} \\
      \hline
      multilingual-e5-base & \ding{51} & \ding{51} & \ding{51} & 90.50 & 89.55 & 89.81 & 93.39 & 90.81 \\
      multilingual-e5-large & \ding{51} & \ding{51} & \ding{51} & \textbf{90.22} & 89.95 & \textbf{91.83} & 92.95 & 91.24 \\
      gte-multilingual-base & \ding{51} & \ding{51} & \ding{51} & 89.68 & 89.07 & 90.26 & 92.95 & 90.49 \\
      paraphrase-multilingual-mpnet-base & \ding{51} & \ding{51} & \ding{55} & 87.92 & 87.64 & 88.13 & 91.15 & 88.71 \\
      sentence-transformers/LaBSE & \ding{51} & \ding{51} & \ding{55} & 90.84 & \textbf{90.16} & 91.83 & \textbf{94.40} & \textbf{91.81} \\
      \bottomrule
\end{tabular}
}

\caption{Precision@1 results on the Cross-Lingual Semantic Discrimination (CLSD) datasets derived from the WMT19 and WMT21 German--French test sets. 
For each dataset, distractors are added only in the target language, yielding one evaluation direction per set (DE$\rightarrow$FR or FR$\rightarrow$DE). 
The upper block reports direct cross-lingual evaluation, the lower block evaluation via English pivot translation. 
Columns under \textit{Alignment} indicate whether the model was trained or fine-tuned with cross-lingual, paraphrase, or retrieval objectives. 
\textit{Average} is the mean Precision@1 across all four evaluation sets.}
\label{tab:performance_comparison}
\end{table*}

\section{Experiments}

\subsection{Cross-Lingual Semantic Discrimination}
We propose a task that aims to evaluate multilingual embedding methods based on their capacity to find the closest cross-lingual semantic match amidst large bilingual text collections. The goal is to measure the model's ability to identify a parallel sentence as the most similar to the original sentence in the source language. To accomplish this task, a dataset of parallel sentences is required that will be enriched with four distractor sentences in the target language, all of which are aimed to be semantically similar.

In this paper, we test the hypothesis that LLMs can generate difficult examples that look similar on the surface, but convey different meanings. We prompt  GPT-4 (gpt-4-0613) \citep{openai2024gpt4} to monolingually generate four distractors for each original sentence (see Prompt~\ref{fig:prompt}). The distractors are requested to be structurally and lexically similar, whilst being semantically dissimilar. For each language, we manually examine 200 distractors of the original sentences and find that over 98\% of the distractors meet these criteria. Example sentences are shown in Table~\ref{tab:transformation-examples}.

\paragraph{CLSD Datasets} We generate four datasets and validate samples from each, consisting of both language directions of German--French for the WMT19 \citep{barrault2019findings} and  WMT21 \citep{akhbardeh2021findings} test sets. In the 18,928 generated distractors, we observe that the LLM used different modification strategies such as single polarity inversion, entity replacement or even multiple swaps with word-level Jaccard similarity between originals and distractors at ($\mu = 0.47$, $\sigma = 0.17$). Table~\ref{tab:distractor_jaccard} shows  distractor samples.

\begin{figure}[t]
\begin{tcolorbox}[title={Distractor Generation Prompt},label={prompt},colback=white]\footnotesize
Can you provide me with four tricky sentences (numbered) that look structurally and lexically similar but don't have the same meaning. The sentences should be within similar topics and share commonalities with the original. Answer in \{German/French\}! \newline
\{Original Sentence in German/French\}
\end{tcolorbox}
\captionsetup{labelformat=empty}
\caption{Prompt \arabic{figure}: Distractor generation prompt for GPT-4}
\captionsetup{labelformat=default}
\label{fig:prompt}
\end{figure}

\paragraph{CLSD Evaluation}
To evaluate a multilingual embedding model \(E\), we assess its ability to discriminate between a source sentence \(S\), its parallel target sentence \(T\), and a set of distractors \(D\). For a given source sentence \(S\) and its corresponding translation \(T\) in the target language, along with a set of four distractor sentences \(D\), the model must produce an embedding of \(S\) that is more similar to the embedding of \(T\) than to that of any distractor \(d_i \in D\).

To measure performance of embedding models, we use Precision@1 (P@1), defined as the proportion of samples in which the model ranks the true translation as more similar to the source sentence than the four distractors. For example, in Table~\ref{tab:performance_comparison}, the column WMT19 DE$\rightarrow$FR reports the P@1 of identifying the original French target sentence as more similar than its French distractors.

\paragraph{\textbf{Models Evaluated}}

We examine the following five embedding models available through SentenceTransformers \citep{reimers2019sentence}.

\noindent \textbf{multilingual-e5-base/-large} (M-E5-b/l) \citet{wang2024textembeddingsweaklysupervisedcontrastive,wang2024multilingual} transform XLM-Ro\-BER\-Ta \citep{conneau2020unsupervised} into a bi-encoder through weakly supervised contrastive pre-training with multilingual data and a second step of supervised fine-tuning on retrieval datasets.

\noindent \textbf{gte-multilingual-base} (M-GTE) \citet{zhang-etal-2024-mgte} pre-train a multilingual long-context encoder (8192) and derive a bi-encoder by a process similar to multilingual-e5 that also includes contrastive training against hard negatives.

\noindent \textbf{paraphrase-multilingual-mpnet-base-v2} (M-MP)
\citet{reimers2020making} use knowledge distillation through cosine loss training on parallel sentences between an English paraphrase trained teacher model (paraphrase-mpnet-base-v2) and a multilingual student model (XLM-RoBERTa \citep{conneau2020unsupervised})   to teach the student to represent sentences closely across multiple languages.

\begin{figure*}[t]
\centering
\resizebox{0.96\textwidth}{!}{\includegraphics{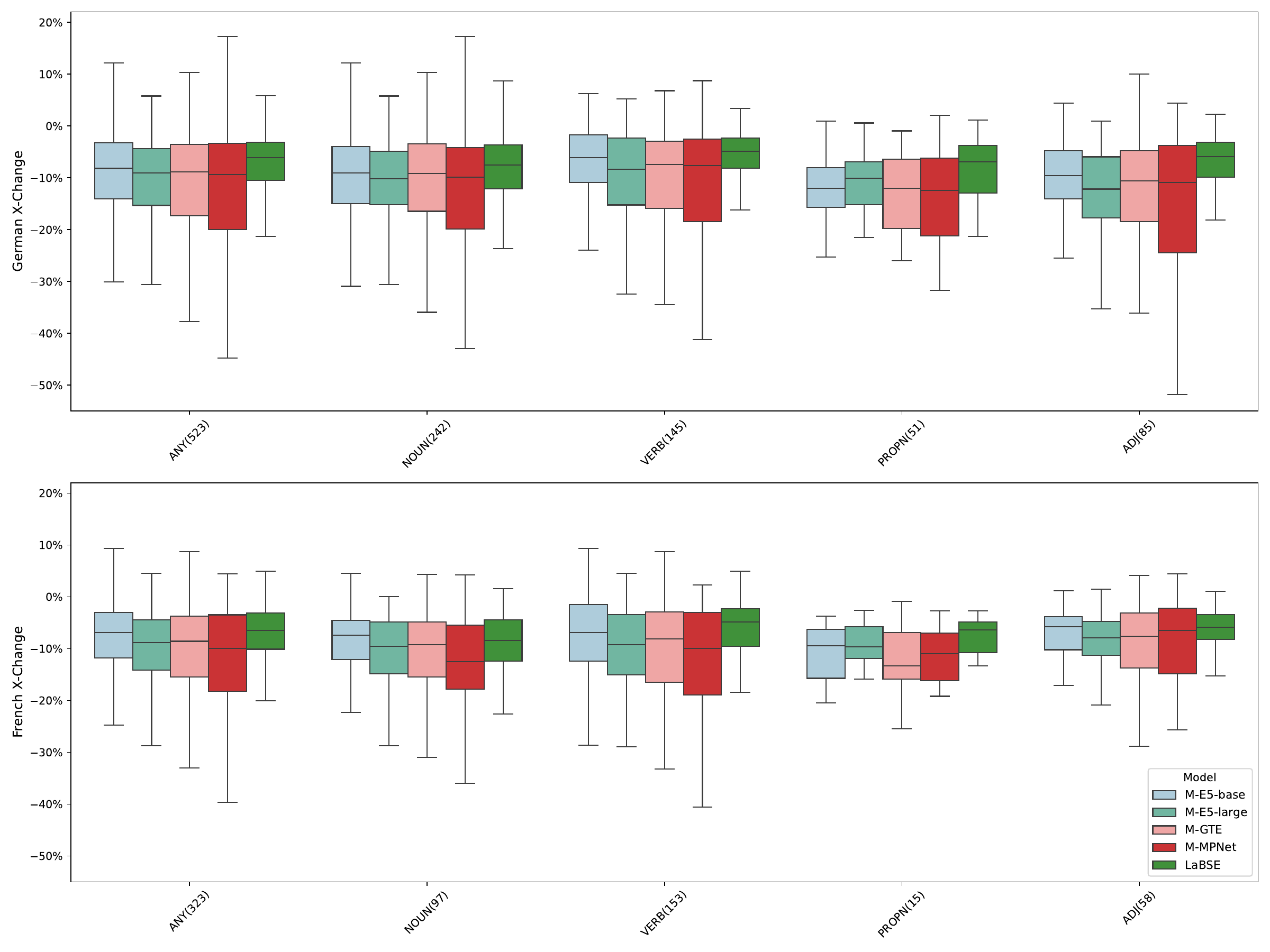}}
\caption{Change in cross-lingual cosine similarity between original and distractor  sentence pairs with exactly one token swapped, grouped by the part of speech of the differing token.}
\label{fig:crosslingual_shift}
\end{figure*}

\noindent \textbf{LaBSE} Language-agnostic BERT Sentence Embedding \citep{feng-etal-2022-language} is trained with translation ranking loss and negative samples.
\subsection{CLSD Results}
The results of the models on the CLSD datasets are shown in  Table~\ref{tab:performance_comparison}.
Regarding the direct cross-lingual evaluation, the Bitext Mining specialist LaBSE is the best model in this task with an average score of 94.43 (P@1), demonstrating its consistent ability to rank the true translation \(T\) higher than any distractor \(d_i \in D\). Another pattern that emerges is that models (M-E5-b/l, M-GTE) that were also trained on English-centric retrieval datasets, such as the MS MARCO \citep{NguyenRSGTMD16}, perform worse within our stricter task.

\paragraph{Pivot Evaluation Setup}
For comparison, we also evaluate models in an English-pivot setting.  For each CLSD dataset, we translate both the source sentence and its candidate set (true translation plus distractors) into English using the M2M 1.2B model \citep{JMLR:v22:20-1307}. The evaluation is then performed monolingually in English. For example, in the WMT19-DE→FR dataset, the German source sentence and all five French candidates are translated into English, and the model must rank the true English translation above its distractors. This yields four additional evaluation setups corresponding to the four CLSD datasets.

In the pivot evaluation, both LaBSE and M-MP show a performance decrease of 2.6/2.8 points compared to direct cross-lingual evaluation. In contrast,  models with weaker performance in direct cross-lingual show only minor improvements, except the M-E5-b model, which gains 3.7 points.

We conduct a qualitative analysis of the 279 samples where LaBSE succeeds in the direct cross-lingual evaluation but fails in the pivot-through-English evaluation. Our analysis reveals that, in most cases, language-specific terms are translated into broader and common English words, reducing the distinction between the original texts and the distractors. Less frequently, we observe omitted details in the translations, hallucinations, and, rarely, poor quality translations.

\subsection{Fine-grained similarity analysis}
How do minor surface changes affect the semantic similarity between sentence pairs? 
Can linguistic analysis provide a more precise characterisation of the changes?
To address these questions, we pick original-distractor pairs \((T, d_i)\) in which the distractor \(d_i\) differs from \(T\) by only a single word. This subset includes 523 German pairs (5.5\%) and 324 French pairs (3.4\%). Using Stanza (v 1.8) \citep{qi2020stanza} and further validation by a linguist annotator, we assign part-of-speech tags to the differing token. 
We measure  the cross-lingual semantic as the difference between \(\cos(E(S),E(T))\) and \(\cos(E(S),E(d_i))\).  To improve comparability in our analysis across different models, we normalize\footnote{The average cosine similarity difference between parallel and unrelated cross-lingual sentence pairs in the original WMT19/21  datasets is the normalization factor in Figure~\ref{fig:crosslingual_shift}.} this score by dividing with {$\parallel$$-$$\nparallel _{\text{ DE } \leftrightarrow \text{ FR}}$ for the corresponding model. Formally, we measure:  $$ \frac{(1 - \cos(E(S), E(T))) - (1 - \cos(E(S), E(d)))}{\|-\nparallel _{\text{ DE } \leftrightarrow \text{ FR}}} $$

Figure~\ref{fig:crosslingual_shift} presents model results, showing that the M-MP model experiences the largest overall shifts in cosine similarity when single-word modifications are made.  The column ``ANY'' reflects  changes across all items, facilitating within-model comparisons to determine which linguistically characterized subgroups exhibit more pronounced changes. Notably, LaBSE, the best performing model within CLSD,  represents  single-word changes with relatively uniform similarity shifts across linguistic categories, unlike the other models. Additionally, we observe that replacing proper nouns consistently decreases semantic similarity more than replacing other parts of speech across all languages and models.

\paragraph{\textbf{Mono- and Cross-Lingual Change Correlations}}
How language-agnostic are the base models when dealing with distractors? Does it matter whether the source sentence is presented in the source or target language?
We  measure the monolingual similarity change of the original to the distractor by substituting $E(S)$ with $E(T)$ in the calculation. The corresponding monolingual semantic change plots can be found in Figure~\ref{fig:monolingual_shift} in the Appendix.
%
%
By correlating the results of the cross- and monolingual change for all one-word change cases (i.e.\ ``ANY''):   The correlation between monolingual and cross-lingual changes is  strong for {M-MP} with 0.93 closely followed by M-GTE with 0.85 and LaBSE with 0.82. The M-E5-b/l shows  lower correlation with 0.79 and 0.71 respectively.
However, specific nuances depend on  the language, model, or type of replaced word. For example, M-GTE shows a low correlation of 0.64 for proper nouns between cross- and monolingual evaluations in French, compared to 0.77 for German.

\begin{table}[t]
\resizebox{\columnwidth}{!}{%
\begin{tabular}{c|r!{\vrule width 1pt}r|r|r|r|r}
\toprule
Lev Sim.& $|D_{\text{bin}}|$ 
& M-E5-b & M-E5-l & M-GTE & M-MP & LaBSE \\
\midrule
\multicolumn{7}{c}{\textbf{French}} \\
\midrule
0.9–1.0 & 1120 & 25.9\% & 20.1\% & 26.6\% & 30.4\% & 32.3\% \\
0.8–0.9 & 2138 & 27.1\% & 23.9\% & 31.4\% & 24.2\% & 28.2\% \\
0.7–0.8 & 2951 & 31.6\% & 36.5\% & 26.2\% & 28.5\% & 25.0\% \\
0.6–0.7 & 2394 & 12.6\% & 16.4\% & 12.7\% & 13.0\% & 12.1\% \\
0.3–0.6 &  861 &  2.8\% &  3.1\% &  3.1\% &  3.9\% &  2.4\% \\
\midrule
\multicolumn{7}{c}{\textbf{German}} \\
\midrule
0.9–1.0 & 1523 & 28.4\% & 28.7\% & 30.6\% & 28.9\% & 40.4\% \\
0.8–0.9 & 2534 & 32.9\% & 32.5\% & 33.2\% & 35.1\% & 32.4\% \\
0.7–0.8 & 2914 & 25.4\% & 23.6\% & 22.7\% & 22.2\% & 19.9\% \\
0.6–0.7 & 1890 & 10.3\% & 11.8\% & 10.5\% & 10.3\% &  5.9\% \\
0.3–0.6 &  690 &  3.0\% &  3.4\% &  3.1\% &  3.6\% &  1.5\% \\
\bottomrule
\end{tabular}
}
\caption{Successful distractors for each model, grouped by Levenshtein similarity to the original sentence. 
Values show the fraction of successful distractors (\%). $|D_{\text{bin}}|$ gives the total number of distractors in each bin.}
\label{tab:succ_distractors}
\end{table}

\subsection{Successful Distractors Analysis}
We conduct an analysis that examines the distribution of successful distractors \(d_i \in D\)  with respect to their Levenshtein similarity to the true translation \(T\). Specifically, we binned the distractors per model according to their similarity score and report the relative percentage per bin. Full results are available in Table~\ref{tab:succ_distractors}. 

Across all models, the majority of successful distractors fall into high-similarity bins (especially 0.8–0.99), confirming that minimal changes to the true translation \(T\) can suffice to mislead the models. However, LaBSE stands out with a relatively lower proportion of successful distractors in lower similarity bins, particularly for German. This suggests a greater robustness to more substantial perturbations, potentially explaining its superior performance on our task.

\section{Conclusion}

This paper advances cross-lingual semantic search by identifying gaps in current evaluation practices and introducing adversarial examples for testing multilingual embeddings. In a German--French case study, we contribute four new adversarial cross-lingual evaluation datasets in the news domain. Our comparative experiments highlight that direct cross-lingual and pivot-through-English evaluation each have advantages depending on the embedding model.  Fine-grained analyses further demonstrate how generated distractors reveal differences in the cross-lingual capabilities of multilingual models and show how small perturbations can affect their semantic representations. Future work should extend this approach to additional language pairs and explore more refined adversarial generation methods to improve the evaluation of semantic search models.

\section*{Limitations}
Our pipeline enables cross-lingual evaluation of semantic search for any language pair with available  parallel texts. However, we evaluate only one language pair  and rely on a single LLM for distractor generation, whereas simpler methods might suffice.
Automatic prediction of part-of-speech tags and named entities in our fine-grained analysis can  introduce noise, although manual validation reduces this risk. However, the capabilities of current models in the language of interest should be critically assessed.
Moreover, the evaluation covers only a few models due to space constraints; nevertheless, smaller models could also yield valuable insights. Future work should address these limitations by broadening the range of languages, models, and adversarial generation techniques, ultimately supporting large-scale evaluation  across many languages, including low-resource ones.

\section*{Acknowledgments}

This research is conducted under the project \textit{Impresso -- Media Monitoring of the Past II. Beyond Borders: Connecting Historical Newspapers and Radio}. Impresso is a research project funded by the Swiss National Science Foundation (SNSF 213585) and the Luxembourg National Research Fund (17498891). We would also like to thank our colleagues Michelle Wastl and Juri Opitz for their support with annotations and further discussions.

\bibliography{acl_latex}

\appendix

\section{Appendix}
\label{sec:appendix}

\begin{table}[h!]
\centering
\begin{tabular}{l|r}
\toprule
\textbf{Model Parameter} & \textbf{Value} \\ \midrule
Temperature              & 1.0           \\ \midrule
Top-P                    & 1.0            \\  \bottomrule
\end{tabular}
\caption{Parameters used for GPT-4 via chat completion for the distractor generation. Undefined parameters are the default.}
\label{tab:LLM_params}
\end{table}

\begin{figure*}[t]
\centering
\resizebox{\textwidth}{!}{\includegraphics{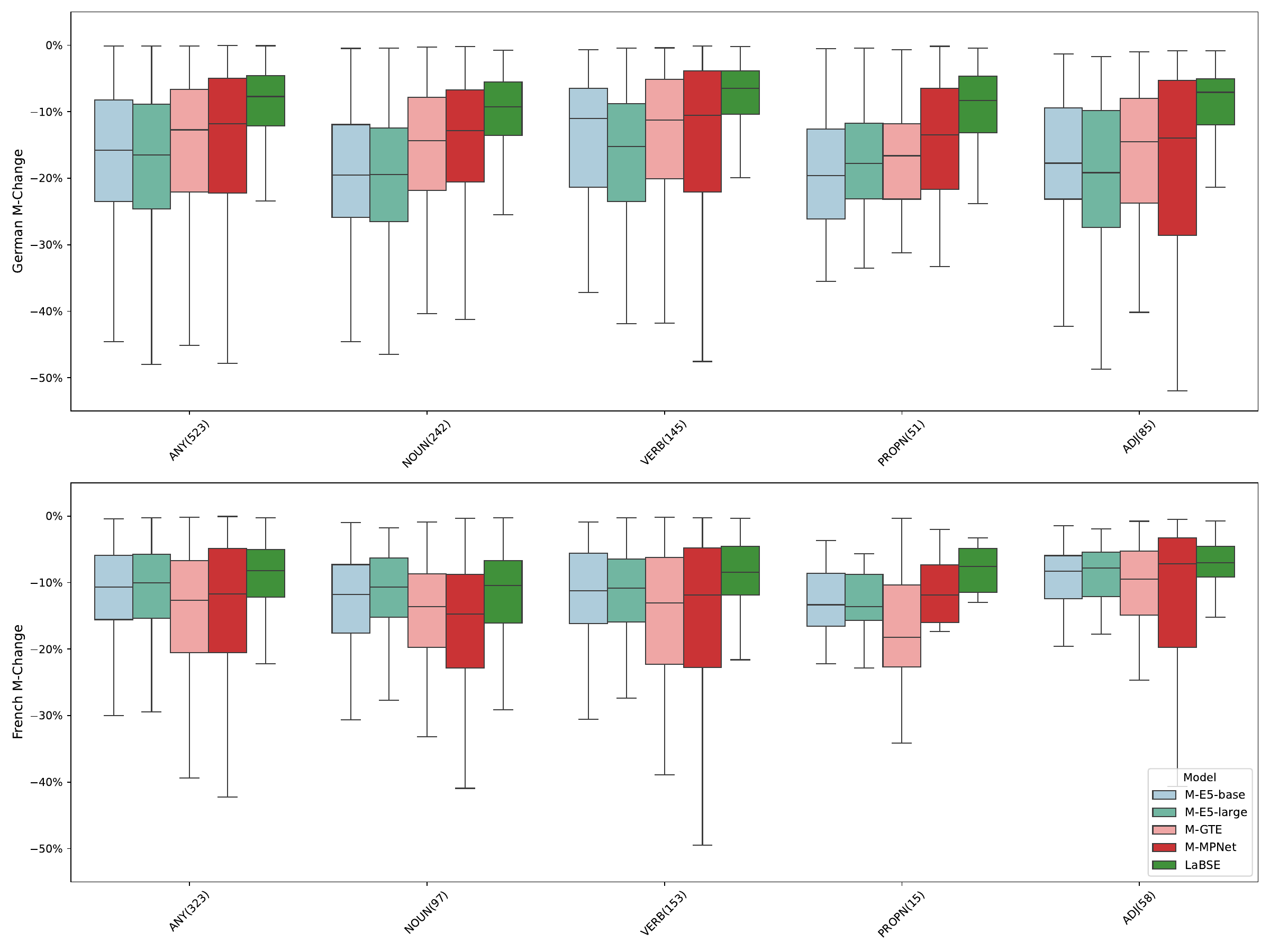}}
\caption{Monolingual cosine similarity change in original-distractor sentence pairs with exactly one token swapped, grouped by the part of speech of the differing token.}
\label{fig:monolingual_shift}
\end{figure*}

\begin{table*}[ht!]
    \centering
    \begin{tabular}{p{12cm} r}
        \toprule
        Distractor & \small InDistr Jaccard Sim. \\
        \midrule
        \textit{Original: PARIS (Reuters) - Die New Yorker Börse schloss am Freitag mit 0,68 \% im Minus und beendete die Woche mit einem allgemeinen Rückgang, da sie von enttäuschenden Ergebnissen, einem Wiederanstieg der Coronavirus-Infektionen und geopolitischen Unsicherheiten beeinträchtigt wurde.} & / \\ \addlinespace
        Adv1: \textcolor{red}{LONDON} (Reuters) - Die \textcolor{red}{Londoner} Börse schloss am \textcolor{red}{Dienstag} mit 0,68 \% im \textcolor{red}{Plus} und \textcolor{red}{startete} die Woche mit \textcolor{red}{allgemeinem Wachstum}, da sie von \textcolor{red}{erfreulichen} Ergebnissen, einem \textcolor{red}{kontinuierlichen Rückgang} der Coronavirus-Infektionen und geopolitischen \textcolor{red}{Stabilitäten angetrieben} wurde. & 0.434 \\ \addlinespace
        Adv2: \textcolor{red}{BERLIN} (Reuters) - Die \textcolor{red}{Berliner} Börse schloss am \textcolor{red}{Donnerstag} mit 0,68 \% im Minus und beendete die Woche mit einem \textcolor{red}{insgesamt flachen Verlauf}, \textcolor{red}{das durch gemischte Ergebnisse}, \textcolor{red}{eine Stagnation} der Coronavirus-Infektionen und geopolitische \textcolor{red}{Spannungen beeinflusst} wurde. & 0.300 \\ \addlinespace
        Adv3: \textcolor{red}{TOKYO} (Reuters) - Die \textcolor{red}{Tokioter} Börse schloss am \textcolor{red}{Mittwoch} mit 0,68 \% im \textcolor{red}{Plus} und \textcolor{red}{begann} \textcolor{red}{den Monat} mit \textcolor{red}{allgemeiner Erholung}, da sie von \textcolor{red}{positiven} Ergebnissen, einem \textcolor{red}{eutlichen Rückgang} der Coronavirus-Infektionen und geopolitischer \textcolor{red}{Sicherheit profitierte}. & 0.359 \\ \addlinespace
        Adv4: \textcolor{red}{MADRID} (Reuters) - Die \textcolor{red}{Madrider} Börse schloss am \textcolor{red}{Montag} mit 0,68 \% im Minus und \textcolor{red}{startete} die Woche mit \textcolor{red}{allgemeinem Rückgang}, da sie von \textcolor{red}{enttäuschenden} Ergebnissen, einem Wiederanstieg der Coronavirus-Infektionen und geopolitischen Unsicherheiten beeinträchtigt wurde. & 0.438 \\ \addlinespace
        \midrule
        \addlinespace
         \textit{Original: Der Nasdaq verzeichnete die schlechteste Woche der letzten vier.} & / \\
        Adv1: Der Nasdaq \textcolor{red}{hat} die \textcolor{red}{beste} Woche der letzten vier \textcolor{red}{verzeichnet}. & 0.630 \\
        Adv2: Der Nasdaq verzeichnete die \textcolor{red}{aktivste} Woche der letzten vier. & 0.623 \\
        Adv3: Der Nasdaq \textcolor{red}{hat} die \textcolor{red}{ruhigste} Woche der letzten vier \textcolor{red}{verzeichnet}. & 0.630 \\
        Adv4: Der Nasdaq verzeichnete die \textcolor{red}{turbulenteste} Woche der letzten vier. & 0.623 \\
        \addlinespace
        \midrule
        \textit{Original: Die Beamten werden im Haushalt für 2021 jedoch nicht vergessen werden.} &  / \\
        Adv1: Die Beamten werden im Haushalt für 2021 jedoch nicht \textcolor{red}{besprochen} werden. & 0.818 \\
        Adv2: Die Beamten werden im Haushalt für 2021 jedoch nicht \textcolor{red}{entlassen} werden. & 0.818 \\
        Adv3: Die Beamten werden im Haushalt für 2021 jedoch nicht \textcolor{red}{befördert} werden. & 0.818 \\
        Adv4: Die Beamten werden im Haushalt für 2021 jedoch nicht \textcolor{red}{belastet} werden. & 0.818 \\
        \bottomrule
        
    \end{tabular}
    \caption{Original-Distractor sets with their Intra-Distractor Jaccard Similarity. Red font indicates modified text.}
    \label{tab:distractor_jaccard}
\end{table*}

\begin{table*}
\centering
\small
\begin{tabular}{>{\raggedright\arraybackslash}p{6.55cm}|>{\raggedright\arraybackslash}p{6.55cm}|>{\centering\arraybackslash}p{1.945cm}}
\toprule
\textbf{Original} & \textbf{Distractor} & \textbf{\small{Tricked Model}} \\ \midrule
\multicolumn{3}{c}{\textbf{WMT19-
DE->FR-CLSD}} \\ \midrule

Kipping au congrès de die Linke sur l'Europe : l'Europe est depuis longtemps un continent d'immigration. & Kipping au congrès de die Linke sur \textcolor{red}{l'Asie  : l'Asie} est depuis longtemps un continent de \textcolor{red}{diversité}. & \textbf{MPNet-M} \\ \midrule

Par exemple, nous allons à présent recevoir un quartier général pour les munitions de l'UE, à Bruxelles, ce que les Britanniques ont jusqu'ici toujours empêché. & 
Par exemple, nous \textcolor{red}{prévoyons de lancer un bureau principal pour les ressources} de l'UE, à Bruxelles, ce que les Français \textcolor{red}{ont toujours refusé jusqu'ici.}
& \textbf{MPNet-M} \\ \midrule

L'appel de Macron: l'Europe est davantage qu'un "projet" & 
L'appel de Macron: l'Europe est \textcolor{red}{plus qu'une "idée".} & \textbf{M-E5-b} \\ \midrule

Trois commissions ont accepté ma candidature, l'une d'elle était de justesse contre. &
Trois \textcolor{red}{comités ont rejeté ma candidature, l'un d'eux était presque en faveur}. & \textbf{M-E5-b} \\ \midrule

\multicolumn{3}{c}{\textbf{WMT19-
FR->DE-CLSD}} \\ \midrule

Würden sie dies tun, gäbe es niemanden in Europa, der sich dagegen wehren würde. &

Würden sie dies tun, gäbe es niemanden in Europa, der sich \textcolor{red}{ dafür aussprechen} würde. & \textbf{MPNet-M} \textbf{M-E5-b} \\
\midrule

Um dem etwas entgegenzusetzen, ist die EU-Kommission auf Kooperation mit den Internetriesen angewiesen. &

Um \textcolor{red}{das zu fördern}, ist die EU-Kommission \textcolor{red}{auf Zusammenarbeit} mit den \textcolor{red}{Internetgiganten} 
angewiesen. & \textbf{MPNet-M} \textbf{M-E5-b} \\ \midrule

Und in der Sozialpolitik sehe ich bei ihr ebenfalls keinen Rechtsschwenk. &
\textcolor{red}{Und bei ihr sehe ich in der Sozialpolitik ebenfalls keinen Linksschwenk.} & \textbf{MPNet-M} \\ \midrule

Bohnentopf, Hähnchen mit Pommes oder Kalbfleisch mit Soße und Kartoffeln: In der Mensa-Eingangshalle der Universität Karlsruhe drängeln sich um die Mittagszeit die jungen Leute. &

\textcolor{red}{Pizzaschnitten, Hähnchen mit Reis oder Rinderbraten mit Gemüse und Kartoffeln: Um die Abendstunde füllt sich der Festsaal der Universität Stuttgart mit hungrigen Studenten.} & \textbf{MPNet-M} \\ \midrule

Nein. Zurückweisungen gibt es ja auch in anderen EU-Ländern. &

Nein, \textcolor{red}{Zuwanderungen} gibt es ja auch in anderen EU-Ländern. & \textbf{M-E5-b} \\  \midrule

Und es ist zugleich die Emanzipation der Linken von ihrem Ex-Vorsitzenden Oskar Lafontaine. &

Und es ist zugleich die \textcolor{red}{Spaltung} der Linken \textcolor{red}{wegen} ihrem Ex-Vorsitzenden Oskar Lafontaine. & \textbf{M-E5-b} \\ 
\bottomrule
\end{tabular}
\caption{Randomly selected examples of successful distractors. The red font indicates modified text.}
\label{tab:succesful_distractors}
\end{table*}

\end{document}